\documentclass[conference]{IEEEtran}
\IEEEoverridecommandlockouts
\usepackage{amsmath}
\usepackage[linesnumbered,ruled,vlined]{algorithm2e}
\usepackage{wasysym}
\usepackage{menukeys} 
\usepackage{graphicx}
\usepackage{textcomp}
\usepackage{xcolor}
\usepackage{url}
\usepackage{booktabs}
\usepackage{balance}
\usepackage[pdftitle={Projecting Robot Navigation Paths: Hardware and Software for Projected AR}]{hyperref}

\begin{document}

\title{Projecting Robot Navigation Paths:\\ Hardware and Software for Projected AR\\
\thanks{\IEEEauthorrefmark{3}Most of this work was completed while Zhao Han was affiliated with the University of Massachusetts Lowell. This work has been supported in part by the Office of Naval Research (N00014-18-1-2503) and the National Science Foundation (IIS-1909864). We also thank 
Brian Flynn at the University of Massachusetts Lowell NERVE Center \cite{nerve} for building the projector support with mechanical springs.}
}

\author{
\IEEEauthorblockN{Zhao Han,\IEEEauthorrefmark{1}\IEEEauthorrefmark{3}, Jenna Parrillo\IEEEauthorrefmark{2}, Alexander Wilkinson\IEEEauthorrefmark{2}, Holly A. Yanco\IEEEauthorrefmark{2} and Tom Williams\IEEEauthorrefmark{1}}
\IEEEauthorblockA{\IEEEauthorrefmark{1}\textit{Department of Computer Science, Colorado School of Mines}, Golden, Colorado, USA 80401\\
Email: zhaohan@mines.edu, twilliams@mines.edu}
\IEEEauthorblockA{\IEEEauthorrefmark{2}\textit{Department of Computer Science, University of Massachusetts Lowell}, Lowell, Massachusetts, USA 01854\\
Email: jenna\_parrillo@student.uml.edu, alexander\_wilkinson@student.uml.edu, holly@cs.uml.edu}}

\maketitle

\begin{abstract}
For mobile robots, mobile manipulators, and autonomous vehicles to safely navigate around populous places such as streets and warehouses, human observers must be able to understand their navigation intent. One way to enable such understanding is by visualizing this intent through projections onto the surrounding environment. But despite the demonstrated effectiveness of such projections, no open codebase with an integrated hardware setup exists. In this work, we detail the empirical evidence for the effectiveness of such directional projections, and share a robot-agnostic implementation of such projections, coded in C++ using the widely-used Robot Operating System (ROS) and rviz. Additionally, we demonstrate a hardware configuration for deploying this software, using a Fetch robot, and briefly summarize a full-scale user study that motivates this configuration. The code, configuration files (roslaunch and rviz files), and documentation are freely available on GitHub at \href{https://github.com/umhan35/arrow_projection}{\textit{https://github.com/umhan35/arrow\_projection}}.
\end{abstract}


\section{Introduction}

Robots are increasingly navigating around populous areas and moving along with people. This is true for service robots found in airports, hotels, restaurants, delivery robots on sidewalks, mobile robots in warehouses, and autonomous cars. 
For these robots to be deployed safely, humans around them must understand their navigation intent.
HRI researchers have traditionally focused on arm movement intent \cite{gielniak2011generating,kwon2018expressing}, exploring eye gaze \cite{moon2014meet,admoni2017social} and other non-verbal means \cite{saunderson2019robots} for stationary robots to better convey such intent. Recently, however, HRI researchers have begun to explore the externalization or visualization of robotic navigation intent (e.g. path plans) as well \cite{coovert2014spatial,szafir2015communicating,chadalavada2015s,watanabe2015communicating,chadalavada2020bi,blenk2021lane}, to enable more understandable robot's navigation behaviors, so as to improve trust and acceptance.

\begin{figure}[t]
\centering
\includegraphics[width=\linewidth]{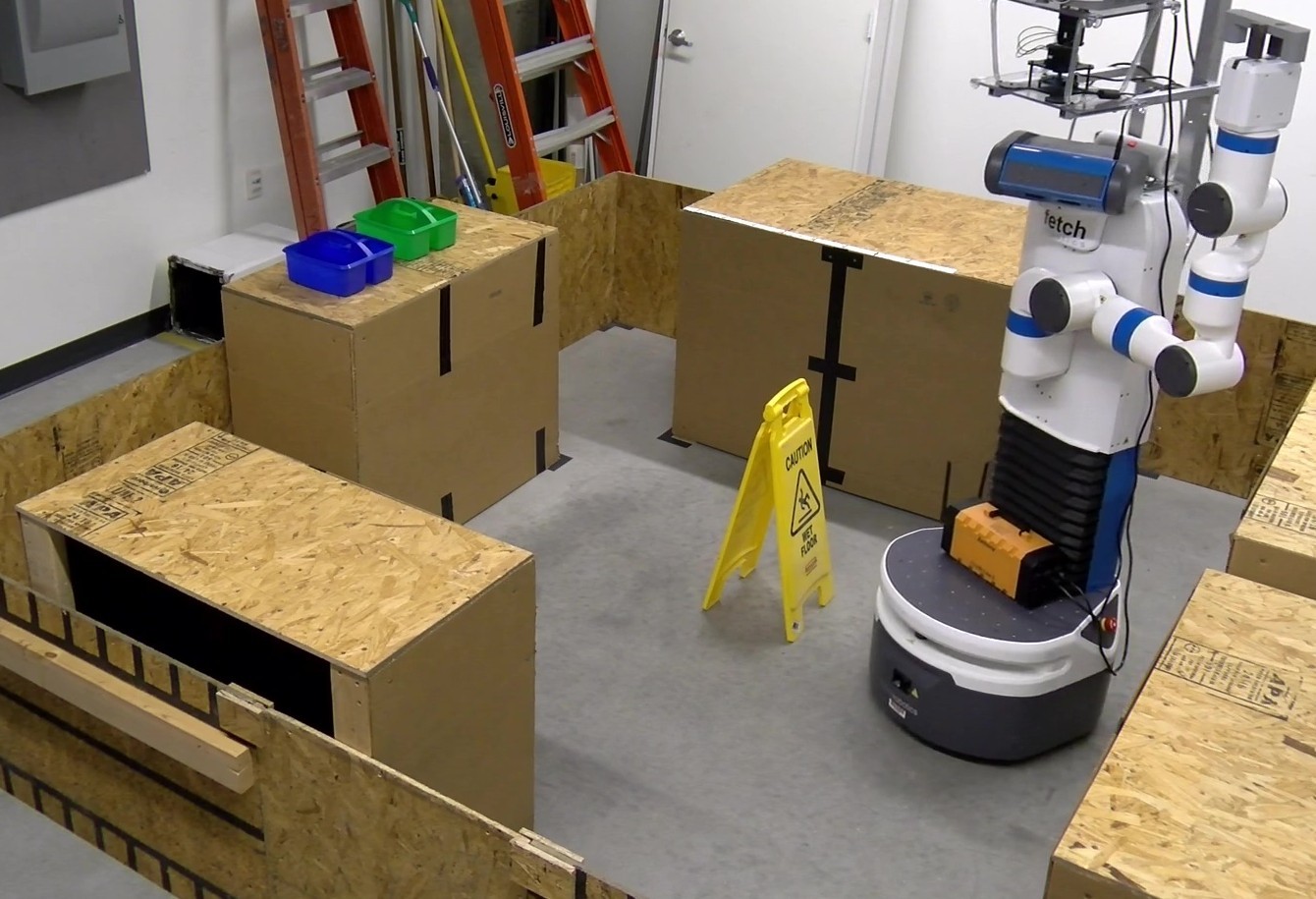}\vspace{1mm}
\includegraphics[width=\linewidth]{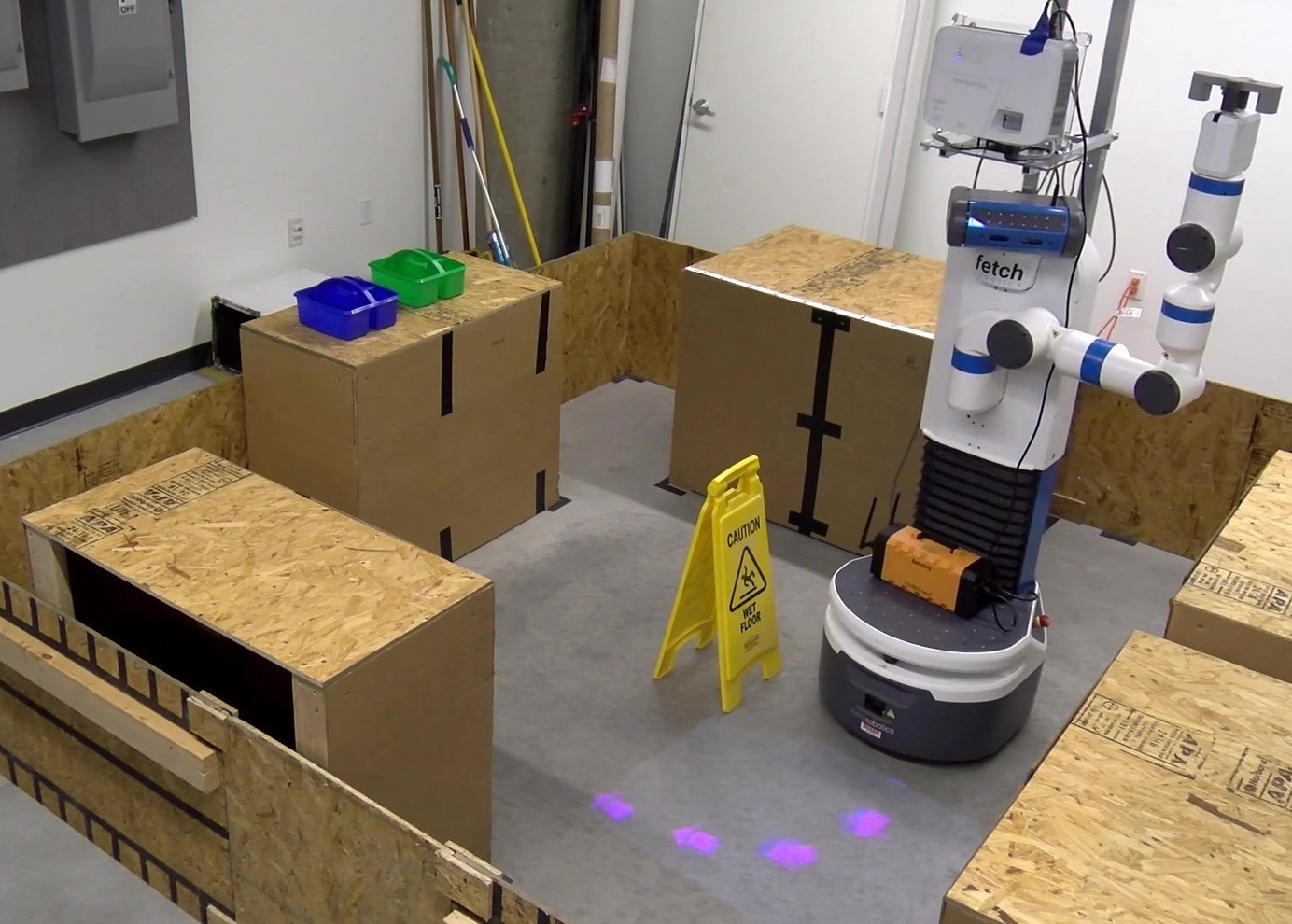}
\caption{A Fetch mobile manipulator navigating around a wet floor sign without (upper) and with (lower) arrow projection (purple). Despite of the effectiveness of arrow projection for revealing navigation path, no public code is readily available. Here we share such code in ROS and provide hardware support. Additional \textbf{videos} are available in \cite{video1,video2}.}
\label{fig:intro}
\end{figure}

One method for conveying navigation intent is through directional projections \cite{coovert2014spatial,chadalavada2015s,watanabe2015communicating,chadalavada2020bi} such as arrow projections \cite{coovert2014spatial,chadalavada2020bi}, where the robot is equipped with a projector and projects directional cues indicating which direction the robot intends to move. These cues can take the forms of lines \cite{chadalavada2015s}, gradient bands \cite{watanabe2015communicating}, or arrows \cite{coovert2014spatial,chadalavada2020bi}. Compared to head-mounted see-through augmented reality \cite{williams2018virtual,williams2018augmented}, projector-based augmented reality does not require interactants to wear special hardware, allowing visualizations to be seen by many observers at once, facilitating usability in group, team, or crowd contexts~\cite{chen2019crowd,sebo2020robots,oliveira2021human}. As we will discuss in Section~\ref{sec:related_work}, a number of human-subjects studies \cite{coovert2014spatial,chadalavada2015s,watanabe2015communicating,chadalavada2020bi} have shown the effectiveness of this approach \cite{watanabe2015communicating,chadalavada2020bi}. 

In this work, we share a practical solution for robotic arrow projections, describe the nature of the implementation of this solution, and provide access to the code for that solution. Additionally, we provide a specific robot and off-the-shelf projector that we have successfully used, as well as a summary of a full-scaled user study to provide more empirical evidence. For the implementation, we used the popular Robot Operating System (ROS) \cite{quigley2009ros} and its visualization tool, rviz \cite{gossow2011interactive}, for rendering the arrows. Using ROS made the implementation robot-agnostic, as most robots used in research and development have ROS support \cite{rosshowcase}. Even non-ROS frameworks \cite{kramer2007development,elkady2012robotics}, and cognitive architectures, such as DIARC~\cite{scheutz2019overview}, typically include bridges to ROS. Using rviz as the rendering engine eliminates the need for practitioners and researchers to learn a tool outside of the ROS ecosystem, such as Unity. Moreover, rviz has a GUI and does not require any computer graphics programming.

Given the wide range of applications of arrow projection on different robots and its proven effectiveness and efficiency, we believe our work is highly relevant and beneficial to the HRI community. The readily available implementation with step-by-step documentation also allows researchers to focus on other interesting and emerging issues under social navigation, e.g., human navigation modeling \cite{guy2010modeling,ratsamee2013social,ferrer2014proactive,che2018avoiding} and robot navigation that obeys social or moral norms \cite{banisetty2021implicit,banisetty2021socially}. We welcome GitHub issues for questions and pull requests.

\section{Related Work on The Effectiveness and Improved Perception of Direction Projection}\label{sec:related_work}

There is significant empirical evidence that directional projections have numerous quantifiable benefits. However, these previous works have not provided publicly accessible software implementations.
Originally proposed to solve the accessibility issue caused by touch screen placement height, Park and Kim~\cite{park2009robots}'s work  was among the first attempts to equip a projector onto a robot, projecting a graphical user interface (GUI) onto the floor so people at different height can see the interface.

Before using arrows, other directional projections were proposed. Inspired by Park and Kim~\cite{park2009robots}, Chadalavada et al.~\cite{chadalavada2015s} proposed projecting a line with a grid onto the ground to indicate navigation path and collision avoidance range of a robot. Even with this primitive geometry element, Chadalavada et al.~\cite{chadalavada2015s}'s work suggests potential for enhanced participant perception of the robot across multiple attributes when projections were used, 
including communication, reliability, predictability, transparency, and situation awareness. Although no statistical tests were ran, their descriptive statistics provide one of the first pieces of evidence for directional projection effectiveness.

Instead of lines, Watanabe et al. proposed projecting a gradient light band in a hallway to show the navigation trajectory to be followed by a robotic wheelchair with a wheelchair user \cite{watanabe2015communicating}. The projection was compared with no projection, and with a projection paired with a screen. Their results suggested that adding projections enhanced comfortability and perceived motion intelligence.  

Furthermore, Watanabe et al.~\cite{watanabe2015communicating} found that when projections were used, nearby walkers chose not to stop and stepped away towards walls earlier. These behavioral changes are further evidenced by Chadalavada et al.~\cite{chadalavada2020bi}, who show that projection onto a shared floor space encourages participants to choose an alternative safer path.

Finally, Coovert et al. used arrow projection to show a robot's path and investigated navigation intention prediction during different periods while the robot was avoiding traffic cones in a hallway \cite{coovert2014spatial}. Their work showed that adding projections led to observable differences at a variety of time scales, as 
participants were typically able to quickly identify the robot's navigation intent. 
Moreover, participants also displayed high confidence in their identifications. 

\begin{figure}[t]
\centering
\includegraphics[height=1in]{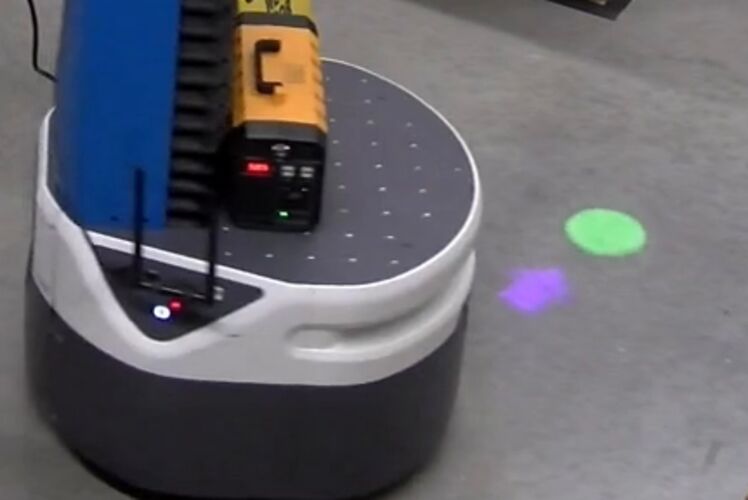}
\includegraphics[height=1in]{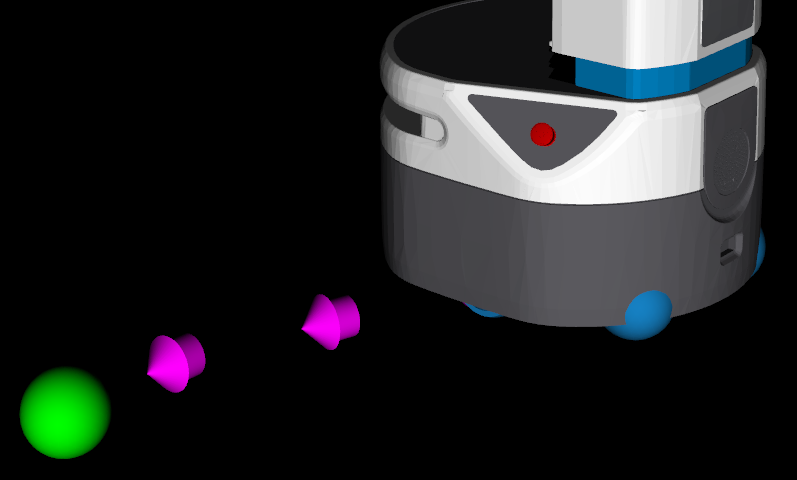}
\caption{\textbf{Left}: A destination circle in green to indicate the end of a navigation path. \textbf{Right}: Arrows with a destination circle rendered in rviz.}
\label{fig:close-up}
\end{figure}

Researchers have also begun to recently compare between some of these previously presented visualizations. Chadalavada et al.~\cite{chadalavada2020bi}, for example, recently compared line and arrow visualizations, and found that arrows are typically preferred compared to lines and blinking arrows. In our own work, we thus made an informed decision to use arrow projections to visualize robot navigation intent. We also added a solid circle to indicate navigation destination (Fig. \ref{fig:close-up} left).

\section{Hardware Setup: Robot, Projector, Power}\label{sec:hardware}

Before we detail the software, we first describe a hardware configuration in which we validated our software. In this configuration , a projector is mounted onto a Fetch robot \cite{wise2016fetch}, as shown in Fig. \ref{fig:intro}. Fetch is a mobile manipulator robot with a single 7-DOF arm mounted onto its chest. Its height ranges from $1.096m - 1.491m$ ($3.596ft-4.892ft$). The projector we used is ViewSonic PA503W \cite{PA503W}, as shown on the top of the images in Fig. \ref{fig:intro}. It used the DLP technology in its lamp, has a brightness of 3,800 ANSI lumens and  22,000:1 contrast ratio, making sure the projection is still legible in indoor environments with lights on. DLP has higher brightness than LCDs because LCDs are transmissive and the heat generated cannot be easily dispatched \cite{hornbeck1997digital}. We chose this projector as our prototype, but other projectors will also work as long as the purchaser makes sure to consider these same three factors, i.e., the lamp technology, ANSI lumens and contrast ratio.

Although we chose the specific robot and projector, our implementation is robot- and projector-agnostic. The only requirement in terms of hardware is the mounting point of a projector must be within the projector's throw distance range to make sure the projection is not blurred. The range should be available on the product specs page \cite{PA503W}.

Because we planned to pan and tilt the projector for tabletop projection, we attached a ScorpionX MX-64 Robot Turret \cite{turret} to actuate the projector. This is not necessary if only used for ground projection, for which the turret is constantly tilted down to the maximum angle. Instead, the projector can be mounted at a fixed position. However, if the actuation of the projector is needed, we do recommend stabilizing the projector, in our case, by using mechanical springs, as shown in Fig. \ref{fig:intro} top.

Another hardware element that might be needed is a portable power station to power the projector as the robot will be untethered. For ViewSonic PA503W, it needs AC power ($100-240V\pm10\%, 50/60Hz$) and has a typical consumption of $260W$, which we found most portable power bank cannot provide (e.g., \cite{nonworkingpowerstation}), leading to projector powered off soon after turned on. We did find a working portable power station with $500W$ AC power outlet \cite{powerstation}. However, if another projector is chosen under the aforementioned three considerations with less power consumption, the power bank is no longer needed.

\section{Implementation in ROS}

Once a projector is mounted onto a robot and pointing towards the front of a robot's base, the projector's pose relative to the robot needs to be integrated into the robot's transform hierarchy using the ``static transform publisher'' node \cite{stp} from the ``tf'' package \cite{tf}. A sample ROS launch file \cite{roslaunch} is provided in ``tf\_publisher.launch''.

Note that to make our account of the implementation beginner-friendly, we added relevant ROS learning resources to references throughout this and following section.

To retrieve the global navigation path, we subscribed to the ROS topic ``/move\_base/TrajectoryPlannerROS/global\_plan'' exposed by the ROS navigation stack \cite{marder2010office,navstack}. The topic encloses a Path message \cite{pathmsg} from the nav\_msgs package \cite{navmsgs}, including poses discretizing the continuous navigation path.

\begin{algorithm}[t]
	\SetAlgoLined
    \DontPrintSemicolon
    \SetKwProg{try}{try}{}{}
    \SetKwProg{catch}{catch}{}{}

	\KwIn{ROS Global Path Poses $P$ \emph{// Unevenly spaced}}
	\KwIn{Double $D$ \emph{// Distance between arrows}}
	\KwIn{Double $\diameter$ \emph{// Destination circle diameter}}

	\KwOut{Array[x,y,z] $P'$}

    $i \leftarrow |P| - 1$ \emph{// From destination to starting point}

    \Repeat{$i > 0$}{

        $p \leftarrow P[i]$,
        $P' \leftarrow P' \cup \{p\}$,
        $i' \leftarrow i$

        \try{}{
            \Repeat{$d < D$ \textbf{or} $(i = |P| - 1$ \textbf{and} $d < D + \diameter$)}{
                $i' \leftarrow i' - 1$

                $p' \leftarrow$ $P[i'].pose.position$ \emph{// ROS quirk}

                $d \leftarrow \sqrt{(p.x - p'.x)^2 + (p.y - p'.y)^2}$

            }
        }
        \catch{Array Out of Bound Exception}{
			\emph{// Done. $i'<0$ now. Line 13 breaks the loop.}
		}

        $i \leftarrow i'$
    }

    \Return $P'$ \emph{// Evenly spaced}

	\caption{Evenly Space Out ROS Nav. Path Points}
	\label{alg:spaceoutnav}
\end{algorithm}

However, our approach uses a voxelized map, an adaptive Monte Carlo localization \cite{fox2002kld}, and an A*-based path planner \cite{konolige2000gradient}, and as such, the poses in the path are not evenly spaced. Because an arrow has a physical size, the arrows may overlap and thus obscure each other if the distance between a pair of poses is too close. We thus created Algorithm \ref{alg:spaceoutnav} to solve this problem. The algorithm also allows specifying a navigation destination circle, as shown in Fig. \ref{fig:close-up} left.

Algorithm \ref{alg:spaceoutnav} iterates each point from the destination to the starting point (Line 1, 2, and 13). The destination point is always the first point in $P'$ (Line 3) because we wanted the destination circle bigger than an arrow. For each point $p$ in the path, it will skip to the point $p'$ when the distance between $p$ and $p'$ is bigger than the desired distance $D$ (Line 9 before ``or''). We stop iterating once the index becomes negative and is accessed (Line 7), caught by the try-catch block (Line 4, 10, and 11), or the distance is shorter than $D$ in the beginning (Line 9 after ``or''). The algorithm is implemented in C++ and available in the get\_sparse\_points function in ``PathProjection.cpp''.

All the logic in this section is coded in the PathProjection ROS node \cite{nodetutorial} in ``path\_projection\_node.cpp'', ``PathProjection.h'', and ``PathProjection.cpp''.

\subsection{Visualizing Arrows in rviz}\label{sec:vizarrow}

Once we get a list of evenly spaced-out points from the ROS navigation path, we create an array of markers \cite{MarkerMsg} as a MarkerArray message \cite{MarkerArrayMsg} in the ``visualization\_msgs'' ROS package \cite{visualizationmsgs}, and publish it so it can be added to rviz \cite{rviz}.

In rviz, an arrow consists of a shaft and head. Under the hood, as seen from its arrow\_marker.cpp \cite{rvizarrowmarkercpp} and arrow.cpp \cite{rvizarrowcpp} source code, rviz uses four parameters for an arrow -- shaft length, shaft diameter, head length, and head diameter.

However, the rviz GUI only allows for specifying three parameters, the scale 3D vector through the Marker message \cite{MarkerMsg} interface, partly because the message was designed to be generic to specify other shapes such as cubes, spheres, or cylinders \cite{MarkerMsg}. The generic design turned out to limit customization of the arrow. To tackle this problem, we spent significant time investigating rviz's source code and directly modified line 108 of arrow\_marker.cpp \cite{rvizarrowmarkercpp}, in which the four parameters of the arrow's shaft and head are set, in . Specifically, we made the shaft's length the same as the arrow's head's. A diff file is available in ``arrow\_marker.cpp.diff''. We hope this can help roboticists to spend more time on their research of interest instead of on arrow customization. 

\subsection{Projecting Arrows in rviz via Projector Lens}

Once the arrow list is published, we can subscribe to the arrow list's ROS topic in order to render the arrows in rviz (Fig. \ref{fig:close-up} right). This renders each arrow at position, $(x,y,0)$, where 0 denotes rendering on the ground plane. In theory, if we can place a virtual camera in rivz at the pose relative to the robot, exactly the same as where the projector is mounted towards the ground (See Fig. \ref{fig:intro}), we can get a 2D image from the camera and it becomes the input to the projector. This is accomplished using the ``rviz camera stream'' plugin~\cite{rvizcamerastream}. We provide a sample rviz configuration file in ``path\_projection.rviz''. Then we used the image view node \cite{imageviewer} to subscribe to that image topic, and made it full screen in Ubuntu's keyboard settings (Toggle full-screen mode).

Finally, to make the projection to the physical world not distorted, the projector lens needs to be calibrated, which we used the pinhole lens model, as seen in \cite{wang2019towards,han2020projection}. A sample CameraInfo message \cite{camerainfo} and a launch file are provided in ``projector\_camera\_info.yaml'' and ``camera\_publisher.launch''.

\section{Evaluation}

In this section, we describe the experimental validation of our arrow projections in the context of human-robot communication during a collaborative mobile manipulation task \cite{han2022past}, where a Fetch robot projected arrows to show its detour path, spheres to indicate obstacles, and projected point clouds \cite{han2020projection} to indicate mis-/recognized objects onto a table (cp. the similar visualization strategy of Reardon et al.~\cite{reardon2018come} in field navigation contexts). In our approach, ground and tabletop projections were compared with the following conditions: physical replay of its past navigation path and manipulation plans (pick and place), and/or speech describing the locations of (a) the objects robot picked/placed and (b) the obstacles around which robots detoured.

This experiment assessed whether our visualizations enabled participants who were temporarily not collocated with the robot to infer the identities and locations of obstacles and of misrecognized or misplaced objects, and the reason for detours. Arrow projections, specifically when paired with sphere projections enabled 93.7\% accuracy for inferring the location of obstacles requiring detours. In contrast, speech-based communication only afforded 83.2\% accuracy.


\begin{figure}[t]
\centering
\includegraphics[height=1.3in]{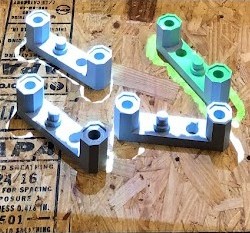}
\includegraphics[height=1.3in]{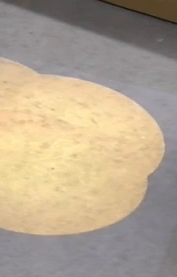}
\includegraphics[height=1.3in]{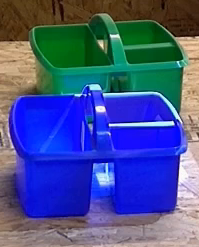}
\caption{Three other visualizations to show the generalizability for our work: point cloud for perceived objects \cite{han2020projection}, multiple spheres for ground obstacle, cube for caddy section. Any rviz visualization can be projected.}
\label{fig:forms}
\end{figure}

As such, this experiment provided empirical evidence for the benefits of our approach. Moreover, we again stress that using our approach, any visualizations that can be added to rviz can be projected (Fig. \ref{fig:forms}), as long as they are in the virtual camera’s view. This includes basic shapes, such as cubes, spheres, and cylinders (visualization\_msgs/Marker \cite{MarkerMsg}) as well as complex objects specified in polygon mesh (MESH\_RESOURCE in Marker \cite{MarkerMsg}; See \cite{meshres} for its usage). However, there are a number of remaining limitations to this approach.

\section{Limitations}

Because the projector must be mounted at a fixed position, it introduces a few limitations. First, the projector must be mounted above the lower value of the projection throw distance range, defined as the distance between projector lens and projection surface. Otherwise, projection onto the ground becomes blurred. This is not a problem for a tall robot, such as the Fetch mobile manipulator, whose height ranges $1.096m - 1.491m$ ($3.596ft-4.892ft$), within the throw distance of ViewSonic PA503W $0.99m-10.98m$ ($3.25ft-36.02ft$). However, mobile robots without upper bodies are often lower than $1m$, such as TurtleBot or PyRobot \cite{murali2019pyrobot}. Using such robots thus requires building a tall structure on the robot, or selecting a projector with a short throw distance.

Another limitation is projection size. This can be clearly seen from Fig. \ref{fig:intro} bottom, the slightly brighter area, which is limited to around $55 \times 98cm^2$ in a $16:9$ ratio. This results in a small projection area, preventing projecting arrows farther along a navigation path, which may lead to problems in large spaces like warehouses. One solution is to mount another projector lower, pointing the ground further away, similar to the headlight of a car, so that the projection can be thrown further to cover a long area, although this may artificially inflate arrow size. Fortunately, because we know the distance between arrows and the projector lens, we can shrink the size of arrows proportionally to distance.

\section{Conclusions}

\begin{figure}[t]
\centering
\includegraphics[width=0.9\linewidth]{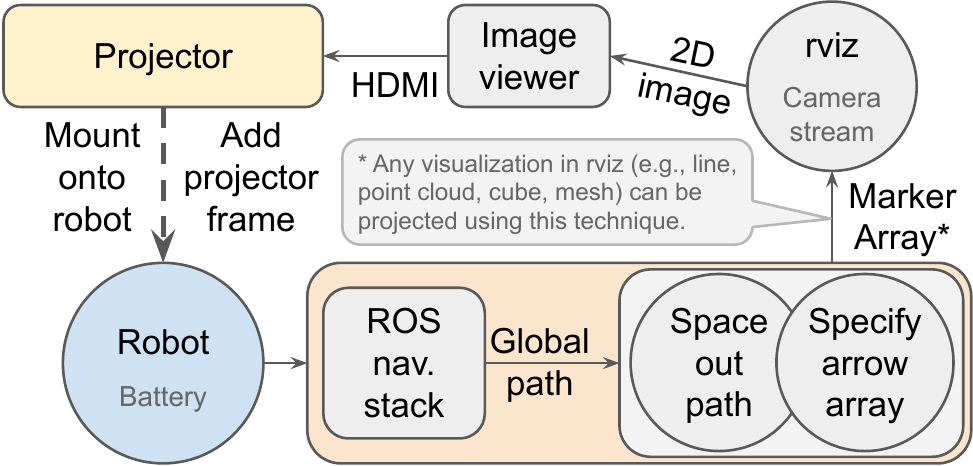}
\caption{High-level representation of the arrow projection implementation, with a note on the generalizability of our technique to other forms of visualizations.}
\label{fig:flow}
\end{figure}

We provided arrow projection software and summarized the demonstrated benefits of this software. As shown in Fig. \ref{fig:flow}, to project arrows, one needs to mount a projector onto a robot with a portable power station, integrate the projector's pose into the robot's transform hierarchy, space out the ROS navigation stack's global path, render the associated arrows in rviz, and finally output to the calibrated projector. Additionally, we also showed the generalization of our technique, i.e., the capability to project any rviz visualizations.

\clearpage
\bibliographystyle{IEEEtran}
\balance
\bibliography{bib}

\clearpage
\end{document}